\pdfoutput=1
\documentclass{article}
\usepackage[final]{colm2026_conference}

\usepackage{microtype}
\usepackage{hyperref}
\usepackage{url}
\usepackage{booktabs}

\usepackage{lineno}

\definecolor{darkblue}{rgb}{0, 0, 0.5}
\hypersetup{colorlinks=true, citecolor=darkblue, linkcolor=darkblue, urlcolor=darkblue}

\usepackage{amsmath,amsfonts,bm}

\def\eqref#1{equation~\ref{#1}}

\def\1{\bm{1}}

\DeclareMathAlphabet{\mathsfit}{\encodingdefault}{\sfdefault}{m}{sl}
\SetMathAlphabet{\mathsfit}{bold}{\encodingdefault}{\sfdefault}{bx}{n}

\usepackage{amsmath}
\usepackage{amsfonts}
\usepackage{amssymb}
\usepackage{wrapfig}
\usepackage{subcaption}
\usepackage{multirow}
 \usepackage{mathtools} 

\usepackage{verbatim}

\usepackage{anyfontsize}

\usepackage{microtype}
\usepackage{graphicx}
\usepackage{booktabs} 
\usepackage{multirow}
\usepackage{amsmath,amssymb}
\usepackage{booktabs}
\usepackage{caption,subcaption}

\usepackage{xcolor}
\definecolor{mygreen}{HTML}{3cb44b}
\definecolor{skyblue}{HTML}{beffff}
\definecolor{lightgreen}{HTML}{90ee90}

\usepackage{color, colortbl}

\definecolor{emerald}{rgb}{0.31, 0.78, 0.37}

\usepackage{tcolorbox}
\usepackage{enumitem}
\usepackage{colortbl}

\usepackage{xcolor}
\definecolor{mygreen}{HTML}{3cb44b}
\colorlet{myyellow}{green!10!orange!90!}
\makeatletter

\usepackage{tikz}
\usetikzlibrary{arrows,shapes,snakes,automata,backgrounds,fit,petri}
\usepackage{adjustbox}

\newcommand{\RN}[1]{%
	\textup{\lowercase\expandafter{\it \romannumeral#1}}%
}
\usepackage{tabu}

\newcommand{\beq}{\vspace{0mm}\begin{equation}}
\newcommand{\eeq}{\vspace{0mm}\end{equation}}
\newcommand{\beqs}{\vspace{0mm}\begin{eqnarray}}
\newcommand{\eeqs}{\vspace{0mm}\end{eqnarray}}
\newcommand{\barr}{\begin{array}}
\newcommand{\earr}{\end{array}}

\usepackage{color, colortbl}
\definecolor{Gray}{gray}{0.93}

\usepackage{lipsum}

\usepackage{pifont}

\usepackage{makecell}

\usepackage{xcolor,amsmath}
\usepackage[linesnumbered,ruled,vlined]{algorithm2e}
\DontPrintSemicolon

\usepackage{xcolor}
\definecolor{mygreen}{HTML}{3cb44b}

\SetKwComment{Comment}{\color{green!50!black}\# }{}

\SetKwProg{Function}{def}{:}{}

\SetKwProg{For}{for}{:}{}
\SetKwProg{If}{if}{:}{}

\usepackage{xcolor}

\definecolor{darkred}{RGB}{140, 21, 21}
\definecolor{lightgray}{gray}{0.7}
\definecolor{orange}{HTML}{F58025}

\title{TEMPURA: Temporal Event Masked Prediction and Understanding for Reasoning in Action}

\author
{\bfseries Jen-Hao Cheng~$^{1}$, Yi-Hao Peng~$^{2}$, Huapeng Zhou~$^{1}$, Vivian Wang~$^{1}$, Huayu Wang~$^{1}$, \\
\bfseries Hsiang-Wei Huang~$^{1}$, Wenhao Chai~$^{3}$, Hou-I Liu~$^{4}$, Kuang-Ming Chen~$^{1}$, Cheng-Yen Yang~$^{1}$, \\
\bfseries Yi-Ling Chen~$^{5}$, Vibhav Vineet~$^{5}$, Qin Cai~$^{ }$, Jenq-Neng Hwang~$^{1}$ \\[0.7em]
{\normalfont\small $^{1}$ University of Washington}\\
{\normalfont\small $^{2}$ Carnegie Mellon University}\\
{\normalfont\small $^{3}$ Princeton University}\\
{\normalfont\small $^{4}$ National Yang Ming Chiao Tung University}\\
{\normalfont\small $^{5}$ Microsoft}\\
[0.7em]
\texttt{Link:
\href{https://andy-cheng.github.io/TEMPURA/}{Project Page}
~|~
\href{https://huggingface.co/datasets/andaba/TEMPURA-VER}{VER Data}~|~
\href{https://github.com/Andy-Cheng/TEMPURA}{Code \& Model}}
\vspace{1em}
}

\begin{document}

\ifcolmsubmission
\linenumbers
\fi

\maketitle
\vspace{-40pt}

\begin{center}
    \centering
    \includegraphics[width=1.0\textwidth]{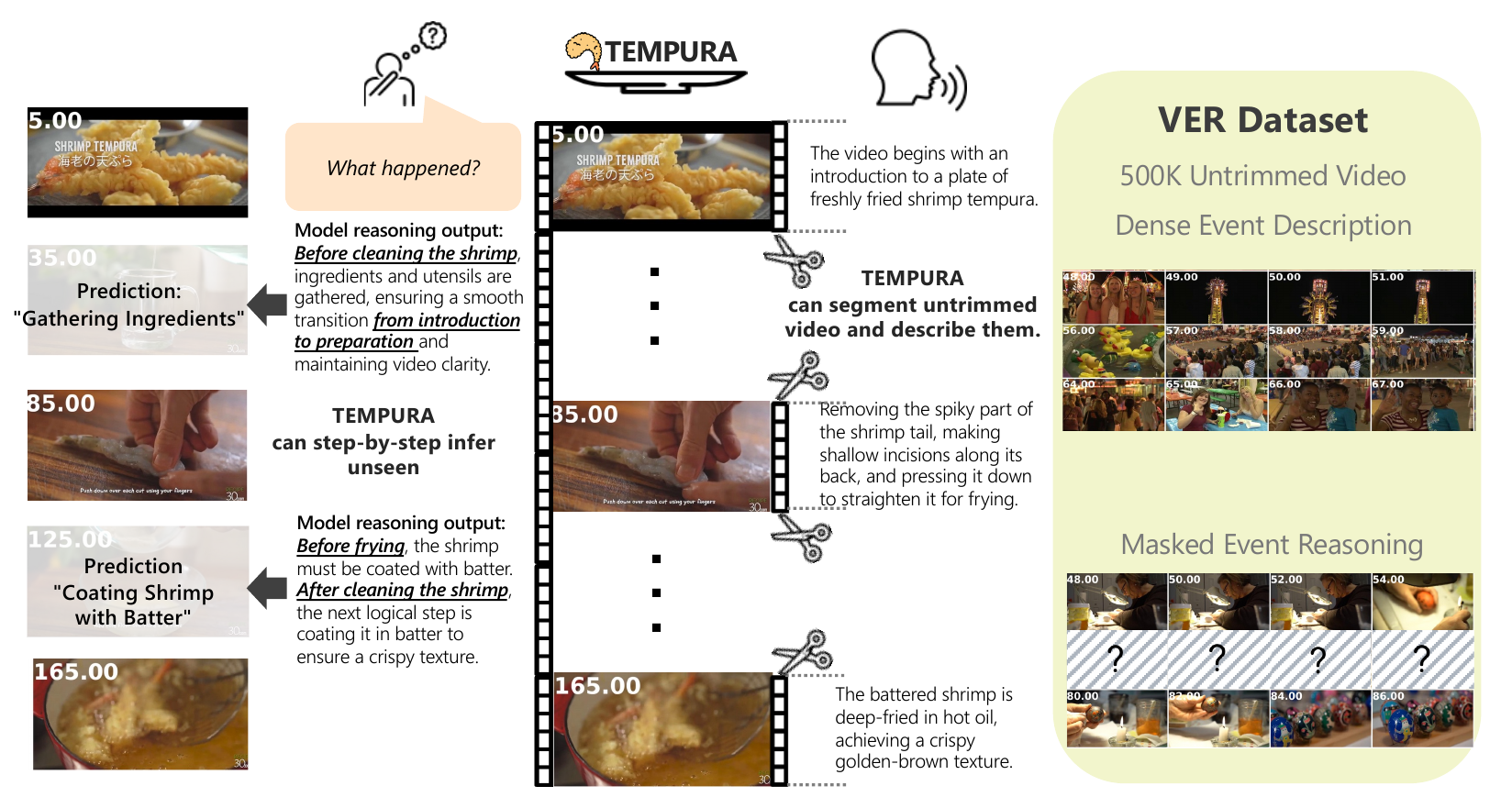}
    \captionof{figure}{
    TEMPURA learns video understanding with a two-stage curriculum. It first reconstructs event structure and causal links by filling missing steps in event sequences. It then segments each video into non-overlapping events and writes dense event captions.}
    \label{fig:intro}
\end{center}

\begin{abstract}
Understanding causal event relationships and achieving fine-grained temporal grounding in videos remain challenging for vision-language models (VLMs). We propose TEMPURA (Temporal Event Masked Prediction and Understanding for Reasoning in Action), a two-stage training framework that enhances the video temporal understanding of VLMs. Inspired by infilling techniques in language modeling, TEMPURA first performs masked event prediction, learning to reconstruct missing events and generate step-by-step causal explanations from dense event annotations. It then learns video segmentation and dense captioning, decomposing videos into non-overlapping events with detailed, timestamp-aligned descriptions. We train TEMPURA on VER, our large-scale dataset of 500K videos annotated with temporally aligned event descriptions and structured reasoning steps. Experiments on video temporal grounding and highlight detection benchmarks show that TEMPURA substantially improves strong base VLMs across model families and scales, confirming that combining event-level reasoning with fine-grained temporal segmentation is an effective recipe for video temporal understanding.
\end{abstract}

\section{Introduction}
\label{sec:intro}
Recent video large multi-modal models~(LMMs)~\citep{llava, llava-onevision, VideoChatGPT, qwenvl25} have extended large language models~(LLMs) with video understanding. Yet fine-grained temporal understanding remains difficult. Recent training pipelines improve video instruction following with curated data or preference optimization~\citep{ln-video, li2025temporal}, but they still do not explicitly teach models to segment videos into coherent events or infer missing causal steps between events.

We address this gap by introducing \textbf{TEMPURA} (\emph{\textbf{\underline{T}}emporal \textbf{\underline{E}}vent \textbf{\underline{M}}asked \textbf{\underline{P}}rediction and \textbf{\underline{U}}nderstanding for \textbf{\underline{R}}easoning in \textbf{\underline{A}}ction}), a two-stage training pipeline that combines event reasoning with dense temporal grounding. In the first stage, TEMPURA masks an event and trains the model to infer the missing event together with a step-by-step causal explanation. Inspired by Fill-in-the-Middle~(FIM)~\citep{FIM, shen2023filmfillinlanguagemodels}, we build this supervision by masking dense event captions and asking a strong LLM to generate a plausible pseudo-event and its reasoning trace. In the second stage, the model learns to segment an untrimmed video into non-overlapping events with explicit start and end timestamps and detailed captions. This stage grounds the event structure learned in stage one and teaches the model to cover the full video without specialized temporal heads. Together, the two stages perform event-centric temporal representation learning: the model first captures how events unfold and connect over time, and then grounds this structure with explicit timestamps.

To support this training recipe, we construct \textbf{VER}, a dataset built through a multi-step event annotation pipeline (Figure~\ref{fig:data_pipeline}). We filter dynamic videos from YT-1B~\citep{yt1b}, categorize them into 10 broad domains with Llama-3-72B~\citep{llama3}, and use GPT-4o~\citep{4o} to segment each video into non-overlapping events from chronologically arranged 1 FPS frame sequences. A temporal coherence filter removes weak event chains before masked event supervision is created. VER contains 500K untrimmed videos, spans 18K hours, and provides 700K training instances with dense event captions and structured reasoning targets.

Under task-zero-shot evaluation, i.e., without fine-tuning on the target benchmarks, TEMPURA improves the Qwen2.5-VL-3B baseline by 6.1 mIoU on Charades-STA and 6.9 HIT@1 on QVHighlights. The pipeline is model-agnostic, and the gains hold across model families and scales: TEMPURA improves InternVL3-2B by 13.9 mIoU, Qwen2.5-VL-7B by 2.8 mIoU and 14.1 HIT@1, and InternVL3-8B by 10.3 mIoU and 23.7 HIT@1. When used as a pre-training stage before task-specific fine-tuning, the improvements grow further to 47.7 mIoU on Charades-STA and 75.1 HIT@1 on QVHighlights. These gains come without sacrificing broader capability: all four TEMPURA-trained models maintain their general video understanding performance on Video-MME~\citep{videomme}. Ablations show that masked event prediction is most effective when it precedes dense captioning: on its own it acts as event-centric temporal pretraining rather than a grounding objective, and applying it first outperforms both dense captioning alone and the reversed order, supporting our design of event reasoning before event grounding.
Our contributions are threefold:
\begin{itemize}
    \item We develop TEMPURA, a two-stage training pipeline that first learns masked event reasoning and then learns dense event segmentation and captioning for grounded video understanding.
    \item We curate VER, a large-scale dataset of 500K videos and 18K hours with timestamp-aligned event captions and masked-event reasoning supervision across 10 common video categories.
    \item We show that the pipeline is model-agnostic and scalable: it improves Qwen2.5-VL and InternVL3 backbones from 2B to 8B on temporal grounding and highlight detection while preserving general video understanding.
\end{itemize}

\section{Related Work}
\label{sec:related_work}
\subsection{Video Large Multi-modal Models}
Researchers have developed video LMMs for broad video understanding tasks~\citep{zhao2024see,zhao2024hierarchical}. Many methods combine vision foundation models~\citep{siglip,clip} with LLMs~\citep{llama3, cai2024internlm2, qwen25} and rely on token merging or compression to process longer videos~\citep{song2024moviechat,chatunivi,slowfast-llava,wu2024freeva,moviechat_plus,chai2024auroracap}. LLaVA-OneVision and LLaVA-Next-Interleave extend the LLaVA design to image and video settings with strong general performance~\citep{llava, llava-onevision, llava-next-interleave}, while Video-LLaMA adds audio for richer multimodal understanding~\citep{videollama,videollama2}. Recent work shows that many of these models still struggle with temporal order and event localization~\citep{timechat,vtimellm,Trace}. We target this weakness with event-level supervision that preserves temporal structure instead of collapsing it.

\subsection{LLM with Reasoning}
LLMs now solve many multi-step problems through stronger reasoning-oriented training~\citep{deepseekr1}. Parallel work in LMMs improves image reasoning with step-by-step supervision for math, sketch, design, localization and agentic setup~\citep{zhang2024mavismathematicalvisualinstruction,shao2024visualcotadvancingmultimodal,peng2024dreamstruct,vinker2024sketchagent,peng2025morae,doveh2025teaching, peng2025designpref,li2024georeasoner,peng2023slide,wu2024uiclip,peng2026efficient}. Methods such as LLaVA-CoT guide image reasoning through staged decomposition~\citep{llava-cot}. Video reasoning remains less explored because a model must preserve temporal order and causal context across dynamic sequences. TEMPURA transfers structured reasoning supervision to this setting through masked event prediction.

\subsection{Temporal Understanding with LMMs}
Existing video LMMs improve timestamp grounding with adaptive compression, segment alignment, or refined temporal embeddings~\citep{timemaker,vtimellm,Momentor,Trace,timesuite,timechat}. These methods focus on temporal localization or captioning, but they usually do not train a model to infer missing events or explain event-to-event causality. TEMPURA complements temporal grounding with masked event reasoning, which improves video coherence and fine-grained event understanding.
Two recent works are closely related. MECD~\citep{mecd} formulates multi-event causal discovery and infers explicit causal graphs among video events; in contrast, TEMPURA uses masked event prediction only as a pretraining objective for downstream grounding and does not attempt formal causal discovery. What Happens When~\citep{whathappenswhen} shows that video LMMs often rely on language priors rather than visual-temporal evidence when reasoning about event order. TEMPURA mitigates this risk in its second stage by grounding every event description to non-overlapping timestamped segments, which requires alignment with the observed video content.

\section{Method}
\label{sec:method}

\begin{figure*}[t]
\centering
    \includegraphics[width=1\textwidth]{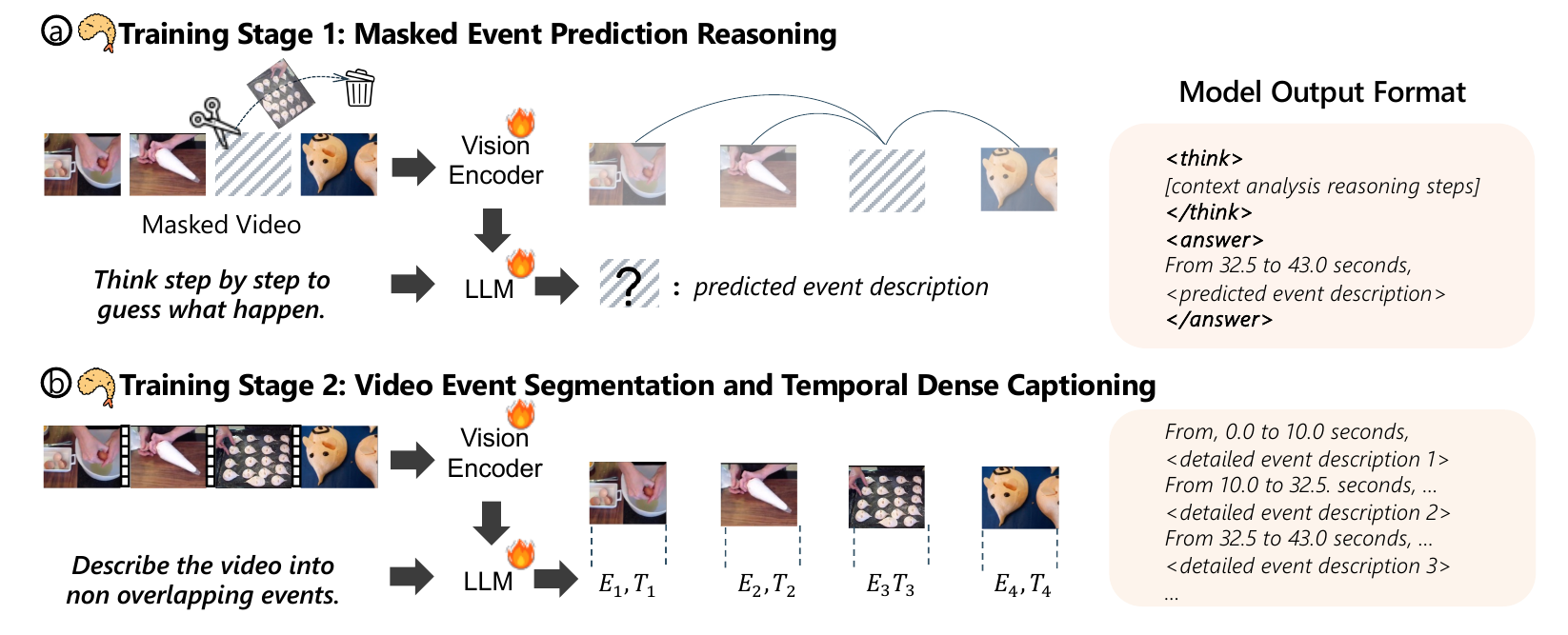}
    \caption{Overview of TEMPURA's two-stage training pipeline. \textbf{(a) Masked Event Prediction Reasoning:} The model infers missing events from masked video context and produces a textual description with step-by-step causal explanations. \textbf{(b) Video Event Segmentation and Temporal Dense Captioning:} The model partitions an untrimmed video into non-overlapping events with start/end timestamps and detailed captions.}
\label{fig:training_stages}
\end{figure*}

To equip video LMMs with the ability to segment videos into meaningful events, establish temporal order, and infer relationships between events, we propose a two-stage training framework. The first stage, Masked Event Prediction, trains the model to infer missing events and reason about event dependencies from video context. The second stage, Video Segmentation and Dense Captioning, teaches the model to decompose videos into temporally grounded event sequences. Together, these stages build a structured understanding of video narratives that generalizes to downstream tasks such as temporal grounding and highlight detection.

\subsection{Masked Event Prediction}
To strengthen video reasoning, we introduce \textbf{Masked Event Prediction}, a training stage that teaches the video LMM to infer a missing event from its surrounding context. Inspired by Fill-in-the-Middle~(FIM)~\citep{FIM, shen2023filmfillinlanguagemodels}, we extend infilling from text to event sequences. Given dense event captions, we mask one event and ask a strong LLM to generate a plausible pseudo-event and a step-by-step explanation that links the missing event to the visible past and future events. Figure~\ref{fig:training_stages}a shows this supervision format.

These pseudo-events and reasoning traces serve as supervised fine-tuning targets. The objective aligns video-based inference with language-based causal reasoning and encourages the model to recover logical event progression from video input alone.

\subsection{Video Segmentation and Dense Captioning}

In the second training stage, we teach the model to break down a video into non-overlapping events and describe each event in detail. As illustrated in Figure~\ref{fig:training_stages}b, the video LMM learns to segment a video into non-overlapping events, each defined by its start and end timestamps, and generates detailed captions for each segment. Unlike Trace~\citep{Trace}, which utilizes extra encoders to model time and saliency scores, we train the model to directly ground all video segments using their enclosing timestamps, leveraging dense video captions from our VER dataset. This eliminates the need for additional parameters, making the video LMM more versatile while ensuring that it learns the structural and temporal progression of videos.

\begin{table*}[t]
    \centering
    \caption{\textbf{Video Dataset Characteristics Comparison across mainstream benchmarks. Note that Charades contains overlapping temporal action annotations.}}
    \label{tab:dataset_stats}
    \small
    \renewcommand{\arraystretch}{1.35}
    \setlength{\tabcolsep}{4pt}
    \resizebox{\textwidth}{!}{
    \begin{tabular}{lcccccc}
        \toprule
        \textbf{Dataset} & \shortstack{\textbf{Video}\\\textbf{Hours}} & \shortstack{\textbf{Events}\\\textbf{per Video}} & \shortstack{\textbf{Events}\\\textbf{per Minute}}
        & \textbf{Coverage} & \shortstack{\textbf{Event}\\\textbf{Details}} & \shortstack{\textbf{Temporal}\\\textbf{Reasoning}} \\
        \midrule
        YouCook2~\citep{youcook2} & 175 & 7.7 & 1.5 & Sparse & $\checkmark$ & $\times$ \\
        Charades~\citep{Charades} & 82.3 & 6.8 & 13.5 & Sparse & $\checkmark$ & $\times$ \\
        ActivityNet Captions~\citep{ActivityNet} & 849 & 3.6 & 1.4 & Sparse & $\times$  & $\times$ \\
        FineVideo~\citep{FineVideo} & 3,425 & 9.9 & 2.1 & Dense & $\checkmark$ & $\times$ \\
        ViTT~\citep{ViTT} & 541 & 7.1 & 1.8 & Sparse & $\checkmark$ & $\times$ \\
        Moment-10M~\citep{Momentor} & 7,260 & 22.5 & 3.3 & Dense & $\checkmark$ & $\times$ \\
        \midrule
        VER~(Ours) & \textbf{18,329} & 10.5 & 6.0 & Dense & $\checkmark$ & $\checkmark$ \\
        \bottomrule
    \end{tabular}
    }
\end{table*}

As summarized in Table~\ref{tab:dataset_stats}, existing datasets lack large-scale timestamp-aligned dense event captions with full video coverage. To support TEMPURA training, we construct Video Event Reasoning (VER), a large-scale dataset of 500K untrimmed videos spanning 18K hours, with non-overlapping events, detailed descriptions, and temporal event reasoning data for masked event prediction. Unlike prior datasets with sparse or partial annotations, VER pairs full-coverage, timestamp-aligned dense captions with masked-event reasoning targets that are retained only when the surrounding event chain passes a temporal coherence check, yielding event-centric supervision.

\begin{figure*}[t]
\centering
    \includegraphics[width=1\textwidth]{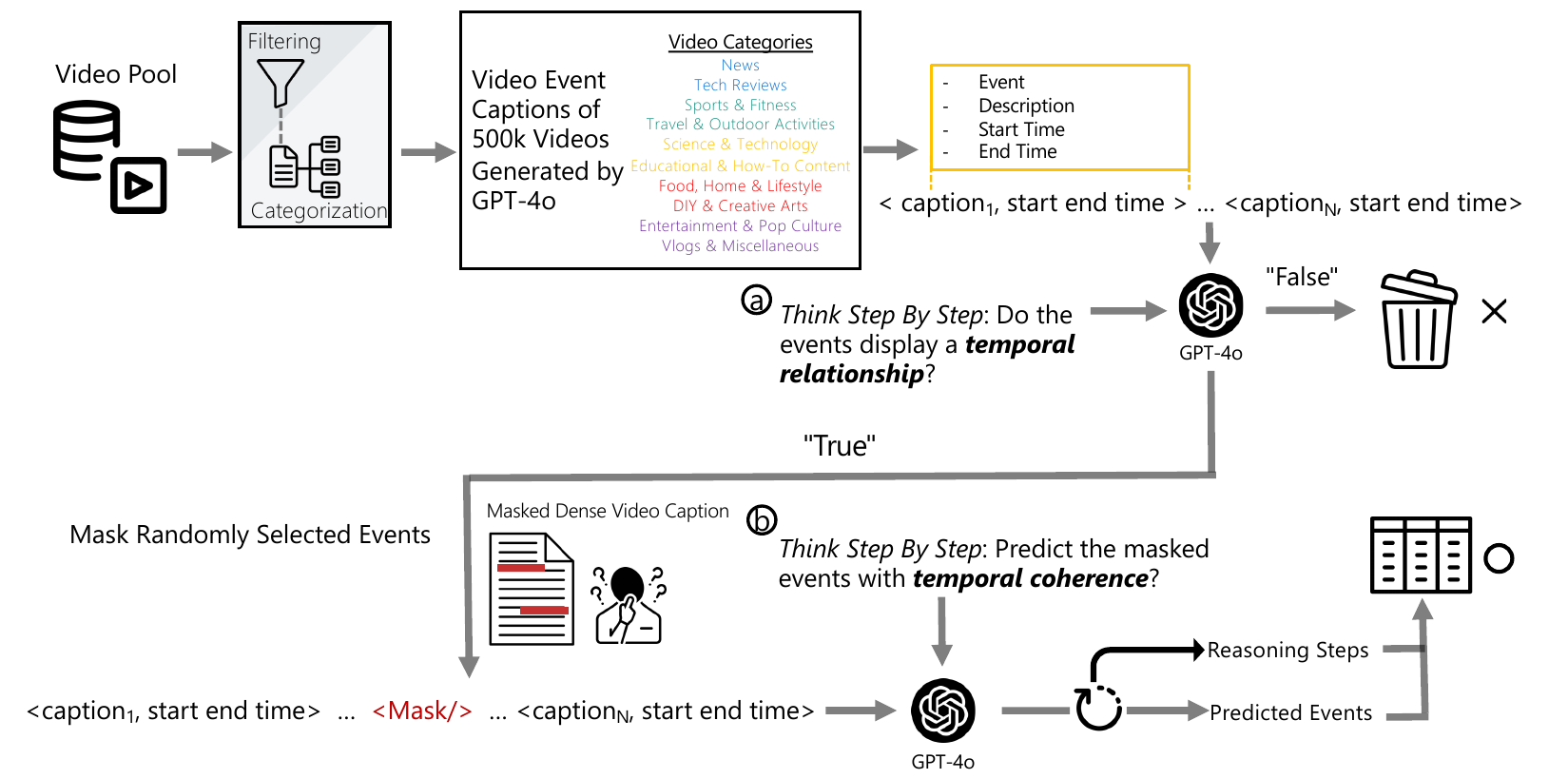}
    \caption{VER data pipeline. We filter and categorize a large video pool, generate dense event captions with GPT-4o, verify temporal coherence, and then create masked-event reasoning targets by masking valid event chains.}
\label{fig:data_pipeline}
\end{figure*}

\begin{figure*}[!t]
\centering
    \includegraphics[width=\textwidth]{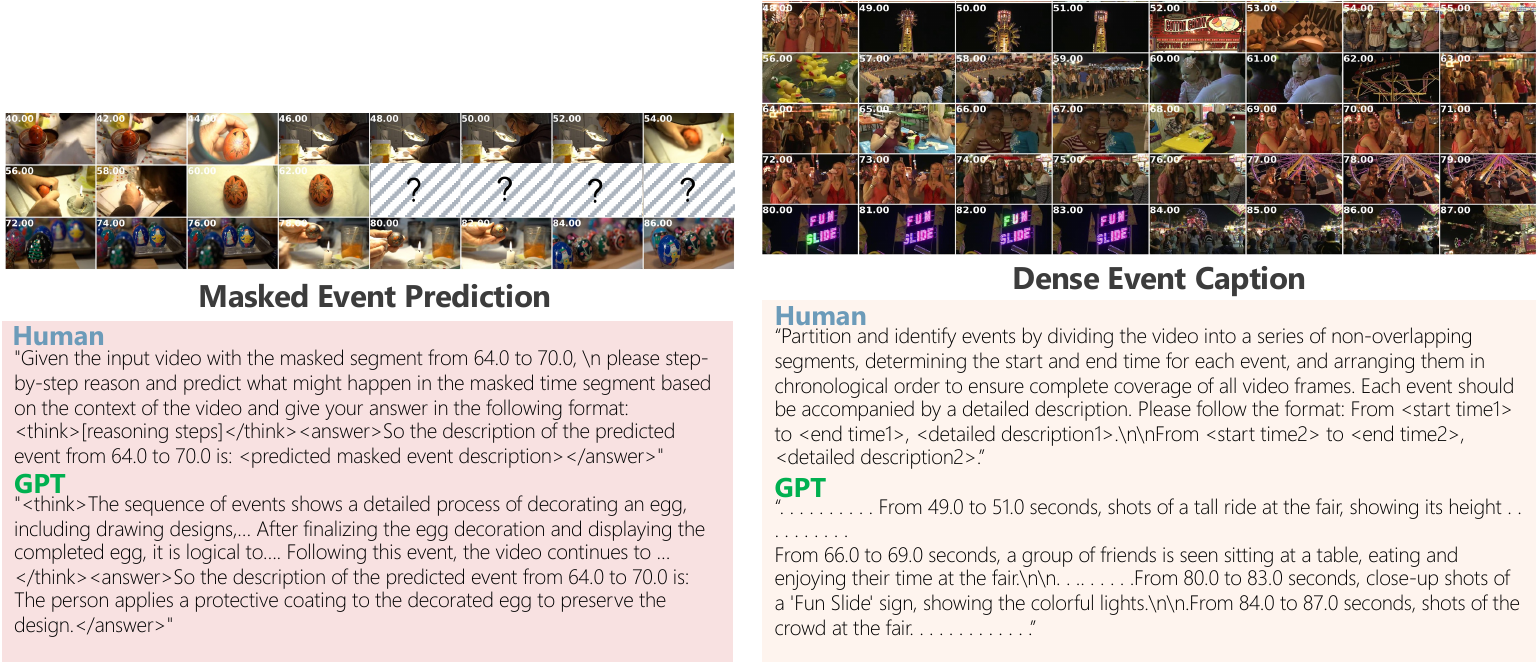}
    \caption{Structured Training Data for Masked Event Prediction and Dense Event Caption}
    \vspace{-10pt}
\label{fig:sft_data_example}
\end{figure*}

\paragraph{Dataset Construction.}
Figure~\ref{fig:data_pipeline} presents our VER data pipeline. Starting from YT-1B~\citep{yt1b}, we remove static videos following \citet{FineVideo} and categorize the remaining videos into 10 predefined categories using Llama-3-72B~\citep{llama3}. To define event boundaries, we sample videos at 1 FPS, arrange frames into chronologically ordered frame sequence images, and apply GPT-4o~\citep{4o} to segment and caption each event. After filtering and alignment, we retain 500K videos with dense event captions, where each event includes a description and start/end timestamps (see Figure~\ref{fig:sft_data_example}).
To create training data for masked event prediction, we randomly mask an event from each dense caption and employ GPT-4o to predict the missing event within the masked time window. To ensure masked events are logically inferable, we filter out videos with uncorrelated event captions by prompting GPT-4o to assess causal relationships via step-by-step reasoning. During training, we align the video LMM’s reasoning with LLM event inference by fine-tuning on these structured reasoning processes, as shown in Figure~\ref{fig:sft_data_example}.

\section{Experiments}
\label{sec:exp}
We fully fine-tuned the base models on the VER dataset following the two-stage TEMPURA training pipeline.

\begin{table*}[t]
\centering
\caption{\textbf{Video Temporal Grounding} on Charades-STA and \textbf{Highlight Detection} on QVHighlights.
Top: prior methods fine-tuned on the benchmark training sets (FT). Middle: prior video LMMs evaluated without benchmark fine-tuning. Bottom: same-backbone pairs comparing each base model with its TEMPURA-trained counterpart under the task-zero-shot setting, i.e., without fine-tuning on the target benchmarks; in the last pair, FT denotes additional fine-tuning on the benchmark training sets.  A dash indicates that the
result was not reported under the corresponding benchmark, metric,
and evaluation setting.}

\small
\renewcommand{\arraystretch}{1.35}
\setlength{\tabcolsep}{4pt}
\resizebox{\textwidth}{!}{
\begin{tabular}{l cccc | cc}
  \toprule
  \textbf{Method}
  & \multicolumn{4}{c|}{\textbf{Charades-STA}}
  & \multicolumn{2}{c}{\textbf{QVHighlights}} \\
  \cmidrule(lr){2-5} \cmidrule(lr){6-7}
  & \textbf{mIoU}
  & \shortstack{\textbf{R@1}\\\textbf{(IoU=0.3)}}
  & \shortstack{\textbf{R@1}\\\textbf{(IoU=0.5)}}
  & \shortstack{\textbf{R@1}\\\textbf{(IoU=0.7)}}
  & \textbf{mAP}
  & \textbf{HIT@1} \\
  \midrule

  QD-DETR (FT)~\citep{qd-detr}
  & - & - & 57.3 & 32.6 & 38.9 & 64.2 \\

  UnLoc-L (FT)~\citep{unloc}
  & - & - & 60.8 & 38.4 & - & - \\

  HawkEye (FT)~\citep{hawkeye}
  & 49.3 & 72.5 & 58.3 & 28.8 & - & - \\

  TimeChat (FT)~\citep{timechat}
  & - & - & 46.7 & 23.7 & 21.7 & 37.9 \\

  VideoChat-T (FT)~\citep{timesuite}
  & - & 79.4 & 67.1 & 43.0 & 27.0 & 55.3 \\

  \midrule

  MovieChat~\citep{song2024moviechat}
  & - & 8.8 & 2.9 & 1.3 & 11.7 & 16.1 \\

  GroundingGPT~\citep{groundgpt}
  & - & - & 29.6 & 11.9 & - & - \\

  VTimeLLM~\citep{vtimellm}
  & 31.2 & 51.0 & 27.5 & 11.4 & - & - \\

  HawkEye~\citep{hawkeye}
  & 33.7 & 50.6 & 31.4 & 14.5 & - & - \\

  TimeChat~\citep{timechat}
  & - & - & 32.2 & 13.4 & 14.5 & 23.9 \\

  TRACE~\citep{Trace}
  & 38.7 & 58.6 & 40.3 & 19.4 & 26.8 & 42.7 \\

  ChatVTG~\citep{qu2024chatvtg}
  & 34.9 & 52.7 & 33.0 & 15.9 & - & - \\

  VideoChat2~\citep{videochat2}
  & - & 9.6 & 3.4 & 1.4 & 13.4 & 18.6 \\

  Momentor~\citep{Momentor}
  & 28.5 & 42.6 & 26.6 & 11.6 & 7.6 & - \\

  Grounded-VideoLLM%
  ~\citep{wang2024groundedvideollmsharpeningfinegrainedtemporal}
  & 36.8 & 54.2 & 36.4 & 19.7 & - & - \\

  \midrule

  Qwen2.5-VL-3B~\citep{qwenvl25}
  & 33.1 & 52.4 & 34.3 & 12.5 & 42.1 & 44.8 \\

  \textbf{TEMPURA-Qwen2.5-VL-3B}
  & \textbf{39.2}
    \textcolor{blue}{\scriptsize(+6.1)}
  & \textbf{63.9}
    \textcolor{blue}{\scriptsize(+11.5)}
  & \textbf{39.3}
    \textcolor{blue}{\scriptsize(+5.0)}
  & \textbf{15.0}
    \textcolor{blue}{\scriptsize(+2.5)}
  & \textbf{48.3}
    \textcolor{blue}{\scriptsize(+6.2)}
  & \textbf{51.7}
    \textcolor{blue}{\scriptsize(+6.9)} \\

  \midrule

  InternVL3-2B~\citep{internvl3}
  & 34.3 & 56.5 & 29.7 & 10.9 & 24.7 & 21.2 \\

  \textbf{TEMPURA-InternVL3-2B}
  & \textbf{48.2}
    \textcolor{blue}{\scriptsize(+13.9)}
  & \textbf{81.8}
    \textcolor{blue}{\scriptsize(+25.3)}
  & \textbf{49.6}
    \textcolor{blue}{\scriptsize(+19.9)}
  & \textbf{17.9}
    \textcolor{blue}{\scriptsize(+7.0)}
  & \textbf{29.3}
    \textcolor{blue}{\scriptsize(+4.6)}
  & \textbf{24.8}
    \textcolor{blue}{\scriptsize(+3.6)} \\

  \midrule

  Qwen2.5-VL-7B~\citep{qwenvl25}
  & 42.7 & 68.5 & 43.7 & 17.3 & 42.2 & 46.2 \\

  \textbf{TEMPURA-Qwen2.5-VL-7B}
  & \textbf{45.5}
    \textcolor{blue}{\scriptsize(+2.8)}
  & \textbf{74.4}
    \textcolor{blue}{\scriptsize(+5.9)}
  & \textbf{47.9}
    \textcolor{blue}{\scriptsize(+4.2)}
  & \textbf{18.8}
    \textcolor{blue}{\scriptsize(+1.5)}
  & \textbf{50.3}
    \textcolor{blue}{\scriptsize(+8.1)}
  & \textbf{60.3}
    \textcolor{blue}{\scriptsize(+14.1)} \\

  \midrule

  InternVL3-8B~\citep{internvl3}
  & 44.6 & 75.6 & 42.5 & 14.4 & 43.5 & 44.5 \\

  \textbf{TEMPURA-InternVL3-8B}
  & \textbf{54.9}
    \textcolor{blue}{\scriptsize(+10.3)}
  & \textbf{91.3}
    \textcolor{blue}{\scriptsize(+15.7)}
  & \textbf{58.7}
    \textcolor{blue}{\scriptsize(+16.2)}
  & \textbf{23.7}
    \textcolor{blue}{\scriptsize(+9.3)}
  & \textbf{53.8}
    \textcolor{blue}{\scriptsize(+10.3)}
  & \textbf{68.2}
    \textcolor{blue}{\scriptsize(+23.7)} \\

  \midrule

  Qwen2.5-VL-3B (FT)~\citep{qwenvl25}
  & 40.3 & 59.0 & 43.1 & 24.7 & 34.7 & 44.4 \\

  \textbf{TEMPURA-Qwen2.5-VL-3B (FT)}
  & \textbf{47.7}
    \textcolor{blue}{\scriptsize(+7.4)}
  & \textbf{65.9}
    \textcolor{blue}{\scriptsize(+6.9)}
  & \textbf{54.9}
    \textcolor{blue}{\scriptsize(+11.8)}
  & \textbf{36.3}
    \textcolor{blue}{\scriptsize(+11.6)}
  & \textbf{57.8}
    \textcolor{blue}{\scriptsize(+23.1)}
  & \textbf{75.1}
    \textcolor{blue}{\scriptsize(+30.7)} \\

  \bottomrule
\end{tabular}
}
\label{tab:video_temp_understanding}
\end{table*}

\subsection{Video Temporal Understanding}
We evaluate TEMPURA on two video temporal understanding benchmarks in a \emph{task-zero-shot} setting, i.e., the model is not fine-tuned on the target benchmark. On Charades-STA~\citep{Charades}, we report mean Intersection over Union (mIoU) and Recall@1 at IoU threshold $\tau$ (denoted R@1@$\tau$). On QVHighlights~\citep{QVHIGHLIGHTS}, we report mean Average Precision (mAP) and HIT@1.

Table~\ref{tab:video_temp_understanding} shows that TEMPURA consistently improves every base model it is applied to. On Qwen2.5-VL-3B, TEMPURA improves Charades-STA by +6.1 mIoU and +5.0 R@1@0.5, and QVHighlights by +6.2 mAP and +6.9 HIT@1. On InternVL3-2B, the gains are even larger: +13.9 mIoU and +19.9 R@1@0.5 on Charades-STA, and +4.6 mAP and +3.6 HIT@1 on QVHighlights. The improvements persist at larger scales: TEMPURA improves Qwen2.5-VL-7B by +2.8 mIoU on Charades-STA and by +8.1 mAP and +14.1 HIT@1 on QVHighlights, and improves InternVL3-8B by +10.3 mIoU, +10.3 mAP, and +23.7 HIT@1. These gains are notable because TEMPURA is evaluated task-zero-shot and uses no target-task fine-tuning, extra temporal heads, or video-specific grounding modules~\citep{groundgpt, vtimellm, timechat, hawkeye, Trace}. We attribute the improvements to event-level supervision that teaches the model to segment videos, infer event transitions, and ground detailed captions in time. TEMPURA also strengthens task-specific fine-tuning: when both models are further fine-tuned with TimeChat's~\citep{timechat} instruction-following data, TEMPURA-Qwen2.5-VL-3B (FT) outperforms the fine-tuned Qwen2.5-VL-3B by +7.4 mIoU and +11.8 R@1@0.5 on Charades-STA, and +23.1 mAP and +30.7 HIT@1 on QVHighlights. Together, the consistent improvements across the Qwen2.5-VL and InternVL3 families from 2B to 8B scales confirm that the benefits stem from the event-centric training signal rather than a specific architecture.

\paragraph{General Video Understanding.}
\begin{table}[!t]
    \centering
    \caption{Video-MME~\citep{videomme} overall accuracy. TEMPURA maintains general video understanding across all base models.}
    \label{tab:videomme}
    \small
    \renewcommand{\arraystretch}{1.0}
    \resizebox{0.52\linewidth}{!}{
    \begin{tabular}{l|ccc}
        \toprule
        \textbf{Backbone} & \textbf{Baseline} & \textbf{TEMPURA} & \textbf{$\Delta$} \\
        \midrule
        Qwen2.5-VL-3B & 54.44 & 55.07 & \textcolor{blue}{+0.63} \\
        Qwen2.5-VL-7B & 57.22 & 57.44 & \textcolor{blue}{+0.22} \\
        InternVL3-2B  & 48.52 & 49.59 & \textcolor{blue}{+1.07} \\
        InternVL3-8B  & 55.19 & 55.41 & \textcolor{blue}{+0.22} \\
        \bottomrule
    \end{tabular}
    }
\end{table}
To test whether event-centric training narrows broader capability, we evaluate all backbones on Video-MME~\citep{videomme}. As shown in Table~\ref{tab:videomme}, TEMPURA maintains overall accuracy on every base model (+0.22 to +1.07), indicating that the gains in temporal grounding and event localization are achieved while preserving general video understanding.

\subsection{Ablation Study}
To study the effectiveness of each component in TEMPURA, we split our ablation analysis into three parts and report mIoU and R@1 on Charades-STA. All ablations use Qwen2.5-VL-3B as the base model.
\paragraph{TEMPURA Training Stage.}

\begin{table}[!t]
    \centering
    \caption{TEMPURA training stages. \textbf{S1}: masked event prediction. \textbf{S2}: event segmentation and temporal captioning.
    }
    \label{tab:training_stage}
    \resizebox{0.58\linewidth}{!}{
    \begin{tabular}{l|ccc}
        \toprule
        \textbf{Training Stages} & \textbf{mIoU} & \textbf{R@1 (IoU=0.3)} & \textbf{R@1 (IoU=0.5)} \\
        \midrule
         S1 & 25.6 & 33.1 & 16.7  \\
         S2 & 38.4 & 59.1 & 32.8  \\
         S2 $\rightarrow$ S1 & 34.0 & 55.6 & 32.6  \\
         \textbf{S1 $\rightarrow$ S2} & \textbf{39.2} & \textbf{63.9}  & \textbf{39.3}  \\
        \bottomrule
    \end{tabular}
    }
\end{table}

Table~\ref{tab:training_stage} shows that masked event prediction is most effective when it serves as the first stage before dense captioning. S1 alone is not a grounding objective: without timestamp-aligned supervision it reaches only 25.6 mIoU, functioning instead as event-centric temporal pretraining that teaches the model to infer missing events and model event continuity. The reverse order, $\text{S2} \rightarrow \text{S1}$, underperforms because the model is not first taught to reason about missing events and event transitions before it learns to localize segments, although we still observe that it extracts facts around the masked time windows and generates longer reasoning steps to predict plausible infills. In contrast, $\text{S1} \rightarrow \text{S2}$ outperforms both S2-only and the reversed curriculum, which supports our design of event reasoning before event grounding.

\paragraph{Dynamic Scene and Relevant Event Filtering.}
\begin{table}[!t]
    \centering
    \caption{Dynamic scene and relevant segment filtering. \textbf{D.S.}: dynamic scene filtering. \textbf{T.R.}: relevant segment filtering.}
    \label{tab:data_filtering}
   \resizebox{0.58\linewidth}{!}{
    \begin{tabular}{l|ccc}
        \toprule
       \textbf{Data Filtering} & \textbf{mIoU} & \textbf{R@1 (IoU=0.3)} & \textbf{R@1 (IoU=0.5)} \\
        \midrule
       No D.S. & 34.8 & 51.3 & 27.4  \\
        D.S. & 38.7~\textcolor{blue}{(+3.9)}  & 63.7~\textcolor{blue}{(+12.4)}& 38.3~\textcolor{blue}{(+10.9)} \\
        \midrule
        No T.R. & 33.0 & 47.3 & 28.3 \\
        T.R. & 37.5~\textcolor{blue}{(+4.5)} & 57.1~\textcolor{blue}{(+9.8)} & 34.8~\textcolor{blue}{(+6.5)} \\
        \bottomrule
    \end{tabular}
    }
\end{table}

Table~\ref{tab:data_filtering} shows that filtering static scenes improves dense captioning supervision, while filtering temporally irrelevant event chains improves masked-event reasoning supervision. Both data filters matter because TEMPURA learns from event progression. Static or weakly related segments dilute the temporal signal and reduce grounding quality.

\begin{table}[!t]
    \centering
    \caption{Performance on Charades-STA under different training frame sampling rates. Inference is fixed at 0.5 FPS for all settings.}
    \label{tab:fps_training}
    \small
    \renewcommand{\arraystretch}{1.0}
    \resizebox{0.58\linewidth}{!}{
    \begin{tabular}{c|ccc}
        \toprule
        \textbf{Training FPS} & \textbf{mIoU} & \textbf{R@1 (IoU=0.5)} & \textbf{R@1 (IoU=0.7)} \\
        \midrule
        0.25 & 37.3 & 36.8 & 14.5 \\
        0.5  & 39.5 & 38.8 & 14.3 \\
        1.0  & 38.9 & 38.4 & 14.2 \\
        \bottomrule
    \end{tabular}
    }
\end{table}

\paragraph{Training Frame Sampling Rate.}
While VER event boundaries are annotated using 1 FPS frame sampling with GPT-4o, the frame sampling rate used during model training is a separate choice. We investigate this by training TEMPURA at 0.25, 0.5, and 1.0 FPS while holding inference fixed at 0.5 FPS. As shown in Table~\ref{tab:fps_training}, TEMPURA performs robustly across all three settings, with mIoU ranging narrowly from 37.3 to 39.5.

\subsection{Discussion}
\label{sec:discussion}
\begin{figure*}[t]
\centering
    \includegraphics[width=1\textwidth]{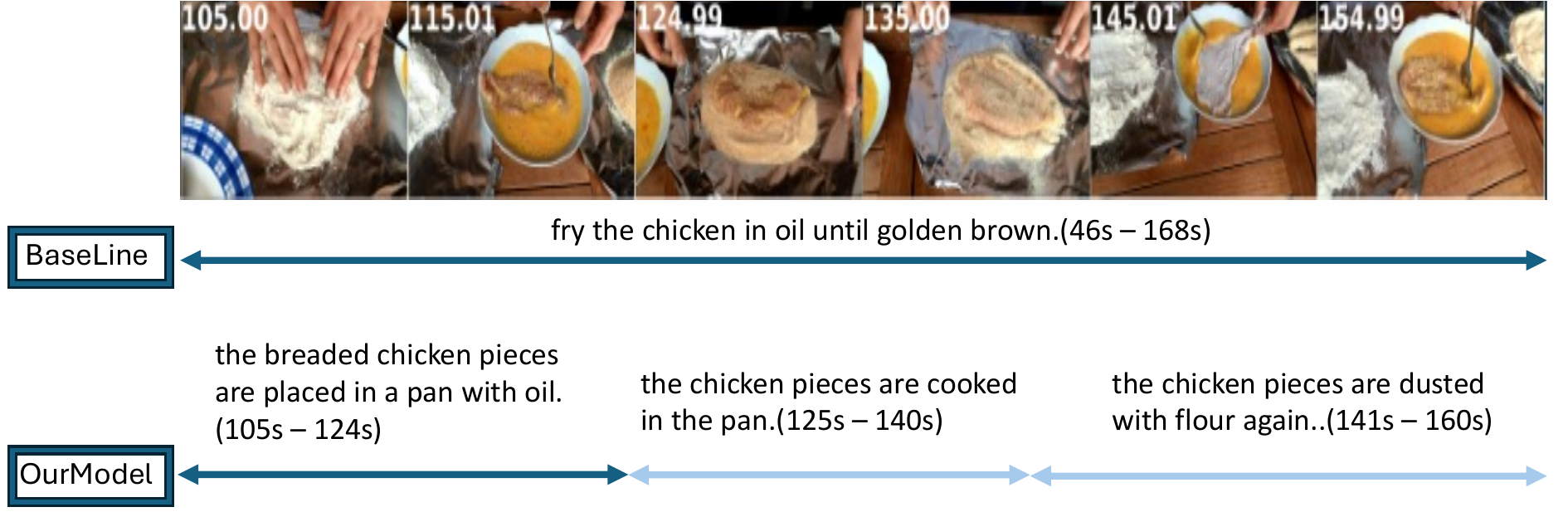}
    \caption{TEMPURA produces finer-grained event partitions than its base model, Qwen2.5-VL, enabling more precise temporally grounded captions.}
\label{fig:Compare_Qwen}
\end{figure*}

The ablations support the central design choice in TEMPURA: event reasoning should come before dense temporal grounding. Stage 1 does not simply expose the model to more captions. It changes the supervision signal. When the model predicts a masked event from visible past and future context, it must represent event transitions, preconditions, and likely consequences. This objective differs from standard dense captioning, where the model can often match local appearance without modeling why one event should follow another. The stage-order ablation in Table~\ref{tab:training_stage} reflects this difference. $\text{S1} \rightarrow \text{S2}$ gives the strongest performance because stage 1 builds an event-level prior over temporal structure, and stage 2 then grounds that prior with explicit boundaries and captions. The reverse order exposes the model to dense event descriptions first, but it does not encourage the model to reason over missing structure before it learns to localize segments.

The downstream gains also align with how TEMPURA decomposes videos. Charades-STA rewards precise alignment between a text query and a short temporal window, while QVHighlights requires the model to rank multiple relevant windows instead of a single best span. Both tasks benefit when the model already represents the video as an ordered event sequence rather than a flat stream of frames. Figure~\ref{fig:Compare_Qwen} and videos on the project page show that TEMPURA infers more coherent segments than Qwen2.5-VL, which leads to better grounded captions for small transitions.

In Table~\ref{tab:video_temp_understanding} fine-tuning (FT) experiments, when TEMPURA is used as a pre-training stage before task-specific fine-tuning, the gains over the baseline grow substantially (+7.4 mIoU on Charades-STA, +30.7 HIT@1 on QVHighlights). This suggests that the event-centric representation learned during TEMPURA's two stages provides a strong foundation that task-specific supervision can build upon. TEMPURA training stages also yields large improvements on InternVL3-2B (+13.9 mIoU) confirms that the benefits are not tied to a specific architecture but stem from the training signal itself.

\section{Limitations and Future Work}
\label{sec:limitation_future_work}
TEMPURA still leaves several open problems. First, event boundaries are inherently ambiguous in videos that contain repetition, gradual state changes, or overlapping sub-actions. A single ``correct'' segmentation may not exist. TEMPURA learns one segmentation that is useful for training, but downstream users may prefer coarser or finer partitions depending on the task. This issue is especially visible in cooking, sports, and vlog-style videos, where small hand motions, object state changes, and camera movement all compete as possible event boundaries. A promising direction is to model boundary uncertainty explicitly rather than forcing one deterministic event chain for every clip.

Second, VER trades annotation scale for annotation purity. GPT-4o provides dense event captions, temporal coherence checks, and masked-event supervision, which makes the dataset large enough to train modern video LMMs. The same design also introduces noise: some captions are too coarse, some boundaries are slightly shifted, and some inferred masked events are only one of several plausible alternatives. Better filtering helps, but it cannot fully solve the problem. A natural next step is to combine automatic annotation with targeted human verification on difficult clips, or to use confidence-weighted training objectives that down-weight uncertain event chains. Another limitation is modality coverage. TEMPURA uses visual supervision only. Audio cues often disambiguate event transitions in instructional, sports, and conversational videos, so multi-modal event supervision may improve the representation substantially.

Third, our evaluation focuses on temporal grounding and highlight detection, which test whether the learned representation transfers to downstream temporal tasks. These benchmarks do not fully measure open-ended reasoning over long videos. Future work should test whether event-centric supervision improves multi-hop video QA, video planning, or interactive retrieval, where a model must compose several events before it answers. The same applies to agents that execute long-horizon tasks alongside people, since an agent must decide when to step in~\citep{yang2026guide, peng2025morae}. It would also be useful to study how TEMPURA interacts with external memory, retrieval augmentation, or planner-style decoders for videos that are much longer than the current training distribution. We view TEMPURA as one part of a broader recipe: learn event continuity, ground events densely in time, and then build higher-level reasoning modules on top of that representation.

\section{Conclusions}
\label{sec:conclusion}

We present TEMPURA, a two-stage training framework that improves video temporal understanding through masked event reasoning and dense event grounding. We also introduce VER, a large-scale dataset that provides both dense event captions and structured reasoning targets. TEMPURA substantially improves base video LMMs on task-zero-shot temporal grounding and highlight detection, yielding consistent gains across the Qwen2.5-VL and InternVL3 families from 2B to 8B scales, and ablations show that reasoning-first training is important for fine-grained video understanding. The model-agnostic pipeline also serves as an effective pre-training stage that amplifies task-specific fine-tuning gains on both Charades-STA and QVHighlights. More broadly, our results suggest that video LMMs benefit from training signals that model events as structured temporal units instead of treating videos as flat frame sequences.

Our findings also point to a practical recipe for future video LMM training. A model first needs supervision that captures continuity across events, because causal ordering is difficult to recover from local appearance alone. Once the model acquires that prior, dense timestamped captions can ground the representation and make it useful for downstream localization and retrieval tasks. The strong fine-tuning results confirm that this event-centric prior transfers well: task-specific supervision builds on the structured representation rather than replacing it. We expect this recipe to complement stronger memory systems, retrieval modules, and planning-oriented decoders rather than compete with them. We hope this perspective motivates future work on scalable event supervision for multimodal reasoning over long-form video.

At a systems level, TEMPURA suggests that better temporal supervision can substitute for some architectural complexity. Several prior approaches improve long-video understanding with specialized temporal heads, additional retrieval components, or benchmark-specific decoding rules. TEMPURA instead improves the base model through an event-centric training signal that is simple to apply at fine-tuning time. This design does not remove the value of specialized components, but it changes where complexity is spent. Rather than adding more inference-time machinery, one can invest in higher-quality event annotations, stronger causal filtering, and denser timestamped supervision. We believe this trade-off is attractive for multimodal research because it scales across tasks and across architectures: once a model learns to represent video as a chain of grounded events, the same representation can support grounding, highlighting, captioning, retrieval, and more open-ended reasoning.

\clearpage
\bibliography{colm2026_conference}
\bibliographystyle{colm2026_conference}
\clearpage

\appendix
\section{Appendix}

\noindent The Appendix is structured as follows:
\begin{itemize}
\setlength{\itemsep}{2pt}
\item VER data creation pipeline and statistics in Section~\ref{supp:annotation}.
\item Qualitative analysis in Section~\ref{supp:qual_analysis}.
\end{itemize}

\section{VER Data Creation Pipeline and Statistics}
\label{supp:annotation}

\begin{figure*}[t]
\centering
    \includegraphics[width=0.65\textwidth]{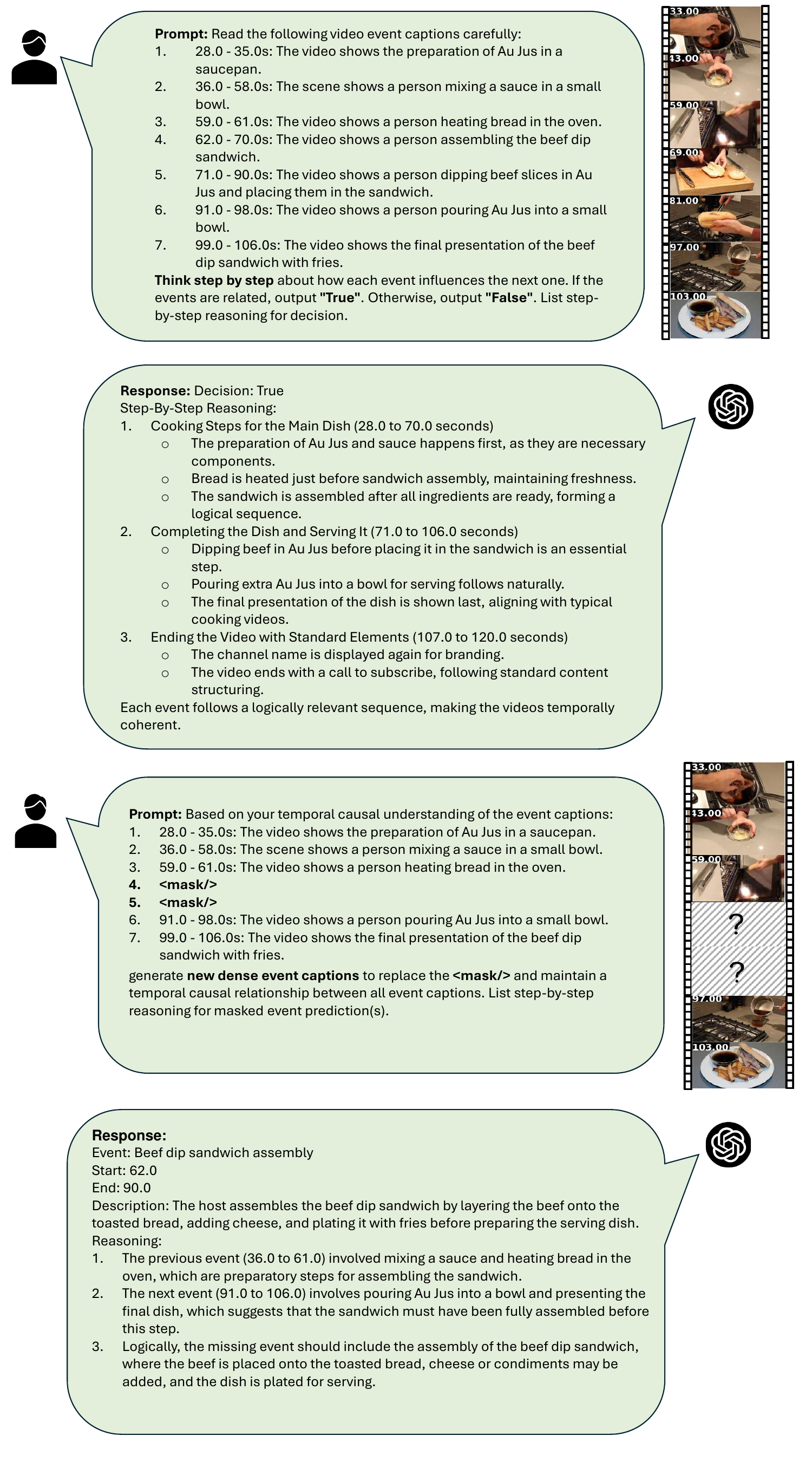}
    \caption{Our mask event prediction data example and generation process.}
\label{fig:data_gen_example}
\end{figure*}

After data filtering, we uniformly sampled video frames and arranged them into frame sequence images.
The example input to GPT-4o is shown on the right of Figure~\ref{fig:data_gen_example}.
We overlayed the timestamp in each image, and combined multiple frame sequence images to obtain a grid-formatted composite image as the input.
These inputs were first temporally segmented by GPT-4o, and we consider these segments as events.
Based on the generated time segments, GPT-4o then generated the descriptions of these events separately.
Next, we used these temporally aligned event descriptions to construct our masked event prediction data.
To create this dataset, we first need to filter out data with weak event correlations.
The bottom half of Figure~\ref{fig:data_gen_example} shows an example of the data pipeline described in Figure~\ref{fig:data_pipeline}.
By inputting the event information into the model in text form, we have it determine whether each correlation is logically valid. After filtering, we obtained 200K masked-event reasoning instances from the 500K dense video captions.

\begin{figure}[htp]
\centering
    \includegraphics[width=.95\textwidth]{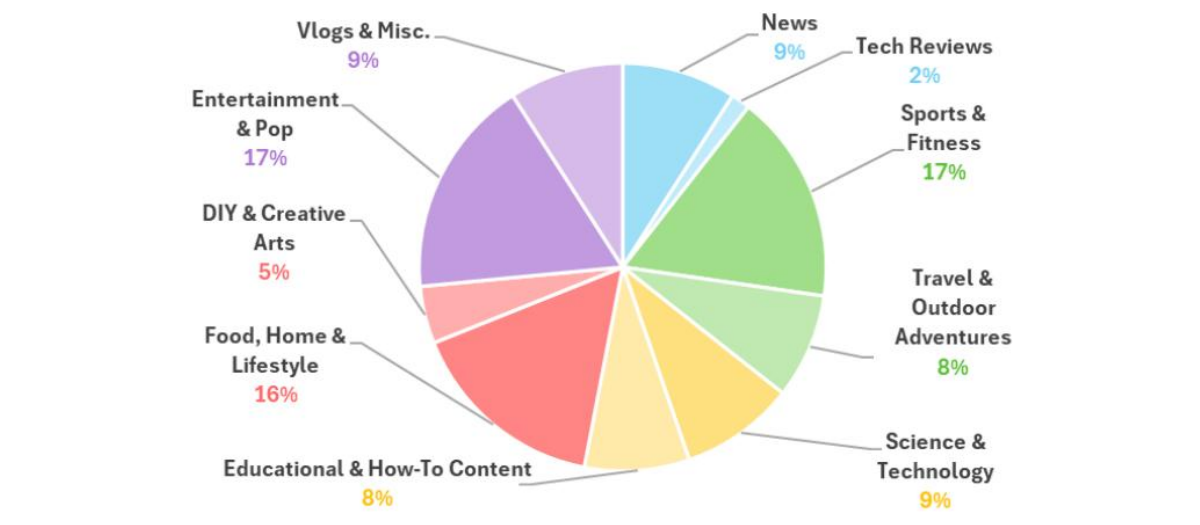}
    \caption{Video Class Distribution}
\label{fig:video_class_dis}
\end{figure}

\begin{figure}[htp]
\centering
    \includegraphics[width=.95\textwidth]{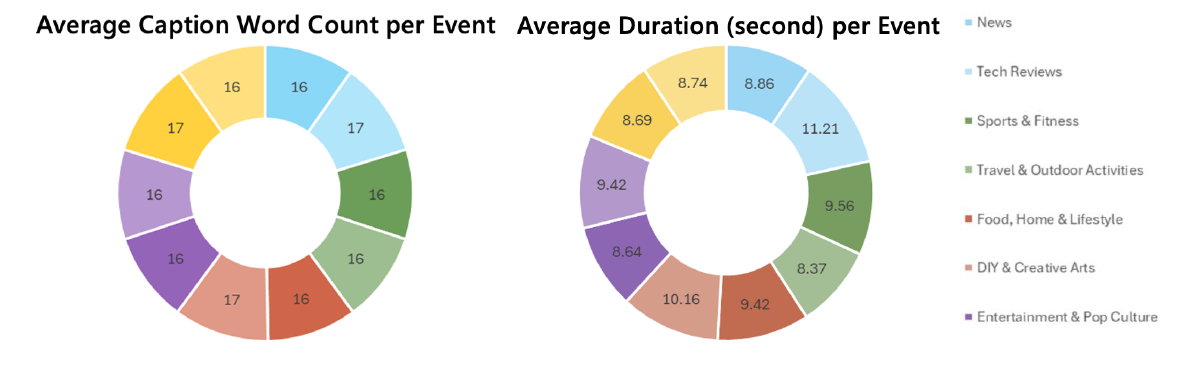}
    \caption{Event Caption Distribution}
\label{fig:event_cap_distribution}
\end{figure}

\begin{figure}[htp]
\centering
    \includegraphics[width=.95\textwidth]{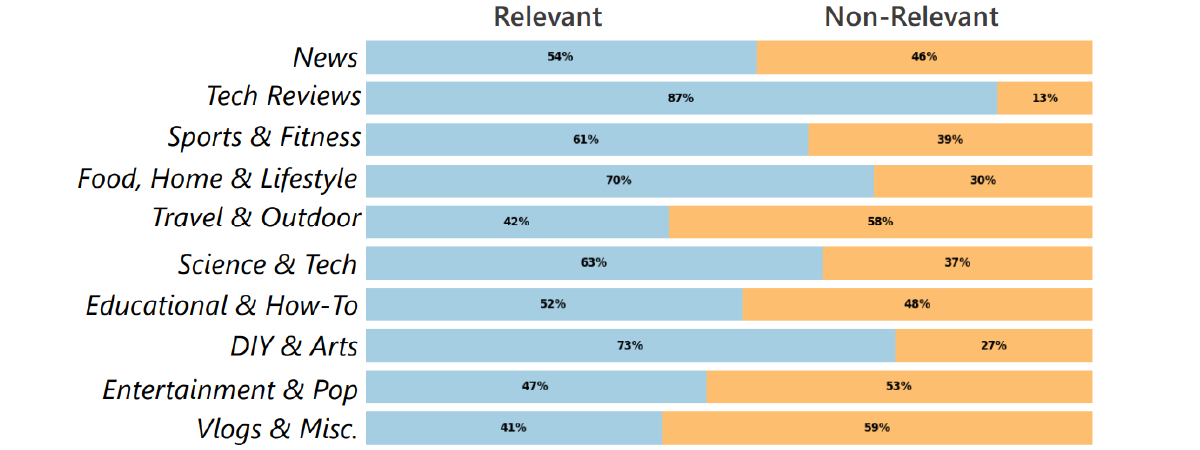}
    \caption{Video Frame Temporal Relevance Distribution}
\label{fig:temp_rel}
\end{figure}

Our dataset contains videos across 10 domains like travel, DIY, tech reviews, etc. (Figure~\ref{fig:video_class_dis}). The average duration and the length of the caption also varied between different domains (Figure~\ref{fig:event_cap_distribution}). Figure~\ref{fig:temp_rel} presents the percentage of videos with temporal relevance in each category.

\section{Qualitative Analysis}
\label{supp:qual_analysis}

\begin{figure*}[t]
\centering
    \includegraphics[width=0.85\textwidth]{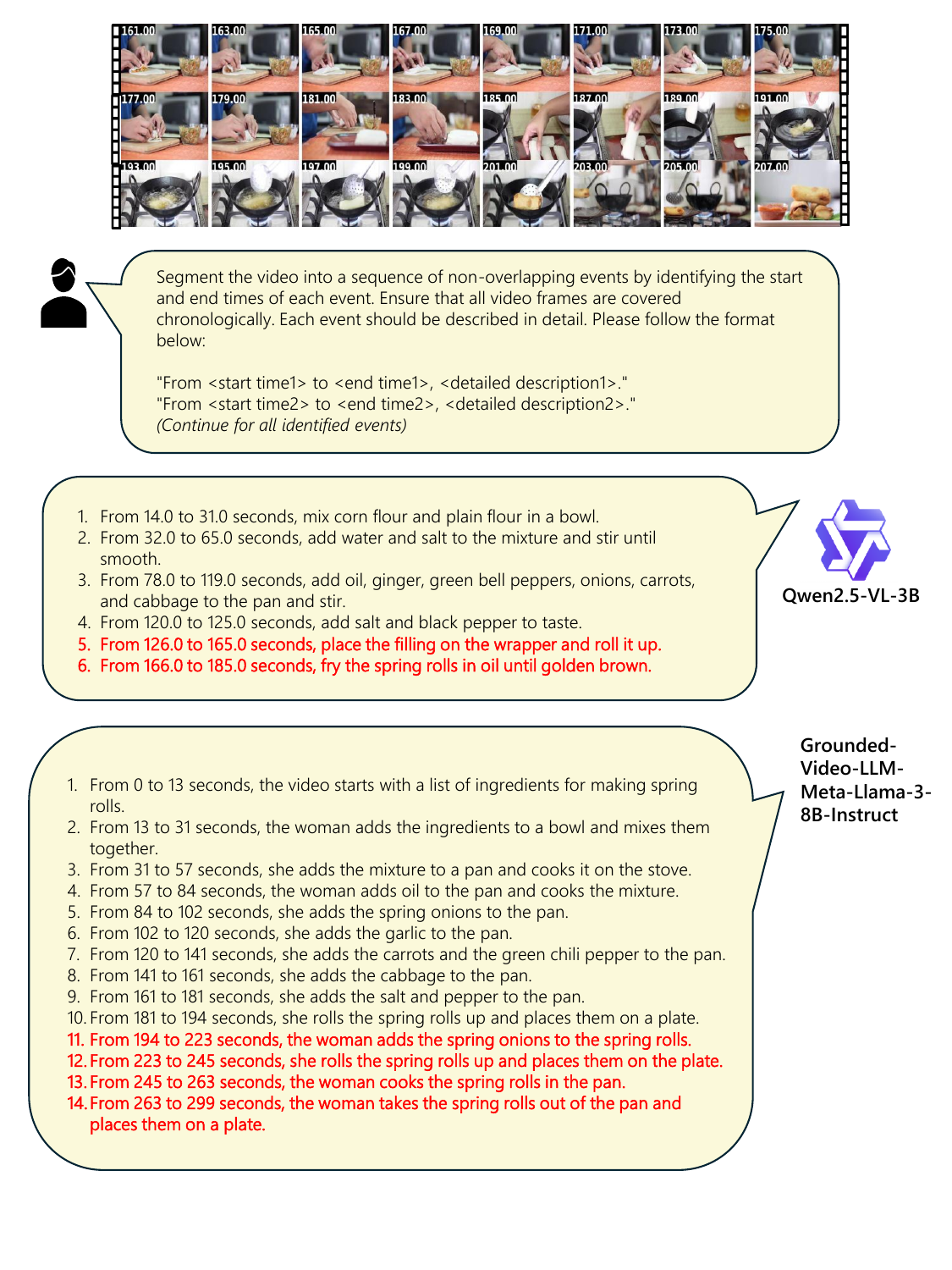}
    \caption{Comparison of \textbf{long video temporal grounding responses} on a cooking tutorial video, generated by baseline models: Qwen2.5-VL-3B~\citep{qwenvl25} and Grounded-Video-LLM-Meta-Llama-3-8B-Instruct~\citep{wang2024groundedvideollmsharpeningfinegrainedtemporal}. Each of the baseline models is prompted to segment videos and describe each of the video segments in detail with correct start and end times. The text highlighted in red indicates incorrect determination of start and end times for frame descriptions.}
\label{fig:temp_rel1}
\end{figure*}

\begin{figure*}[t]
\centering
    \includegraphics[width=0.85\textwidth]{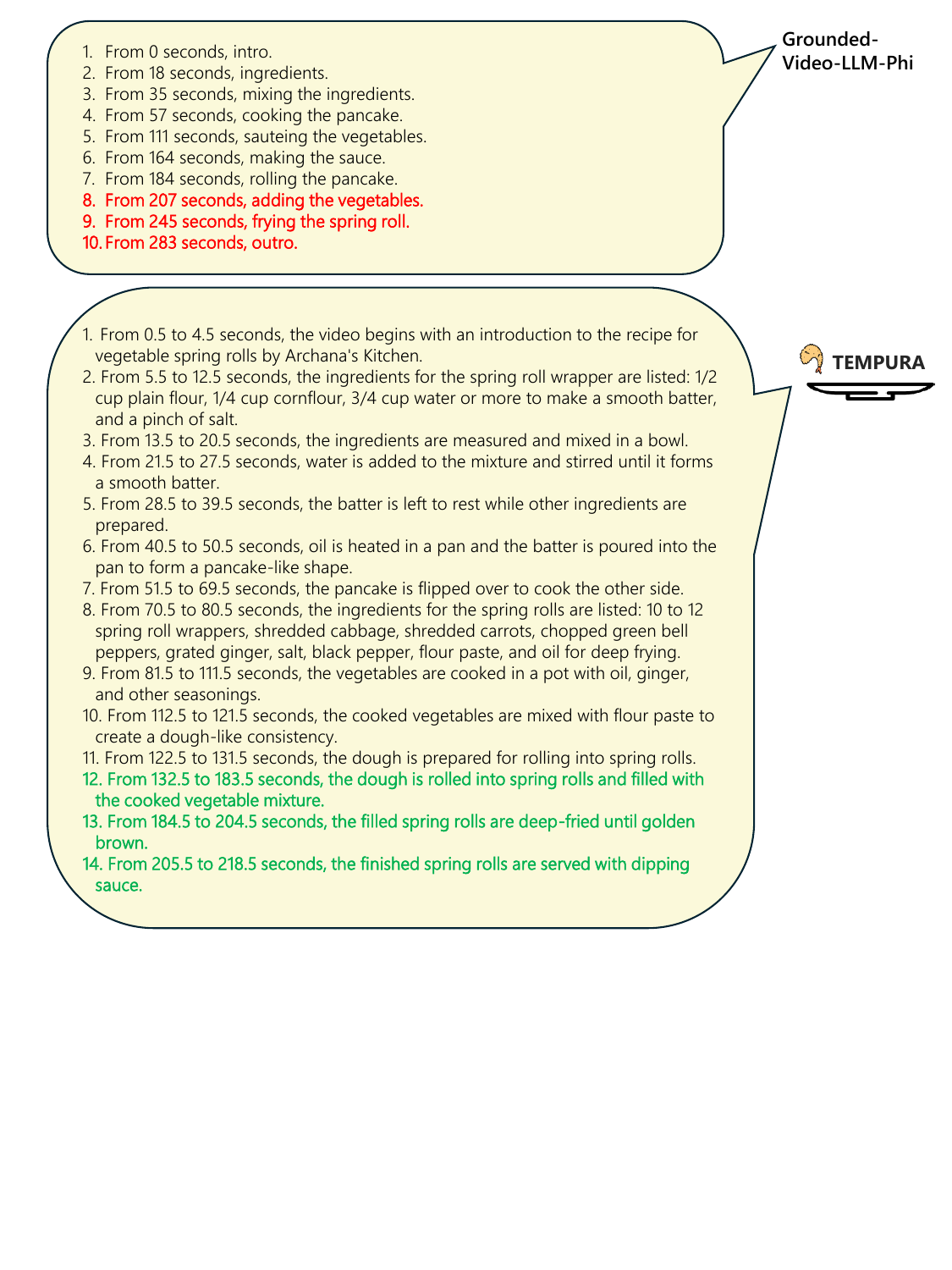}
    \caption{Comparison of \textbf{long video temporal grounding responses} on a cooking tutorial video, generated by TEMPURA (our model) and Grounded-Video-LLM-Phi~\citep{wang2024groundedvideollmsharpeningfinegrainedtemporal} (baseline model). Each of the models is prompted to segment videos and describe each of the video segments in detail with correct start and end times. The text highlighted in red indicates incorrect determination of start/end times and video segment descriptions. The text in green indicates correct determination of start/end times and video segment descriptions.}
\label{fig:temp_rel2}
\end{figure*}

Figure~\ref{fig:temp_rel1} and~\ref{fig:temp_rel2} compare the performance of Qwen2.5-VL-3B~\citep{qwenvl25} (our baseline), Grounded-Video-LLM-Meta-Llama-3-8B-Instruct~\citep{wang2024groundedvideollmsharpeningfinegrainedtemporal}, Grounded-Video-LLM-Phi~\citep{wang2024groundedvideollmsharpeningfinegrainedtemporal}, and TEMPURA (our proposed model) on long video temporal grounding tasks.
The red text highlights errors in timestamp predictions when other models segment videos into fine-grained events and identify their start and end times.
While other models often struggle, especially toward the end of long videos, TEMPURA consistently segments events accurately and assigns precise timestamps.
For instance, the green text shows that TEMPURA correctly identifies a person filling and wrapping spring rolls from 161.00 to 183.00 seconds, followed by placing them into oil for frying from 185.00 to 205.00 seconds.
Additionally, TEMPURA has better performance in producing more fine-grained event captions, as shown by the larger number of event captions produced by our model.

\begin{figure*}[t]
\centering
    \includegraphics[width=0.85\textwidth]{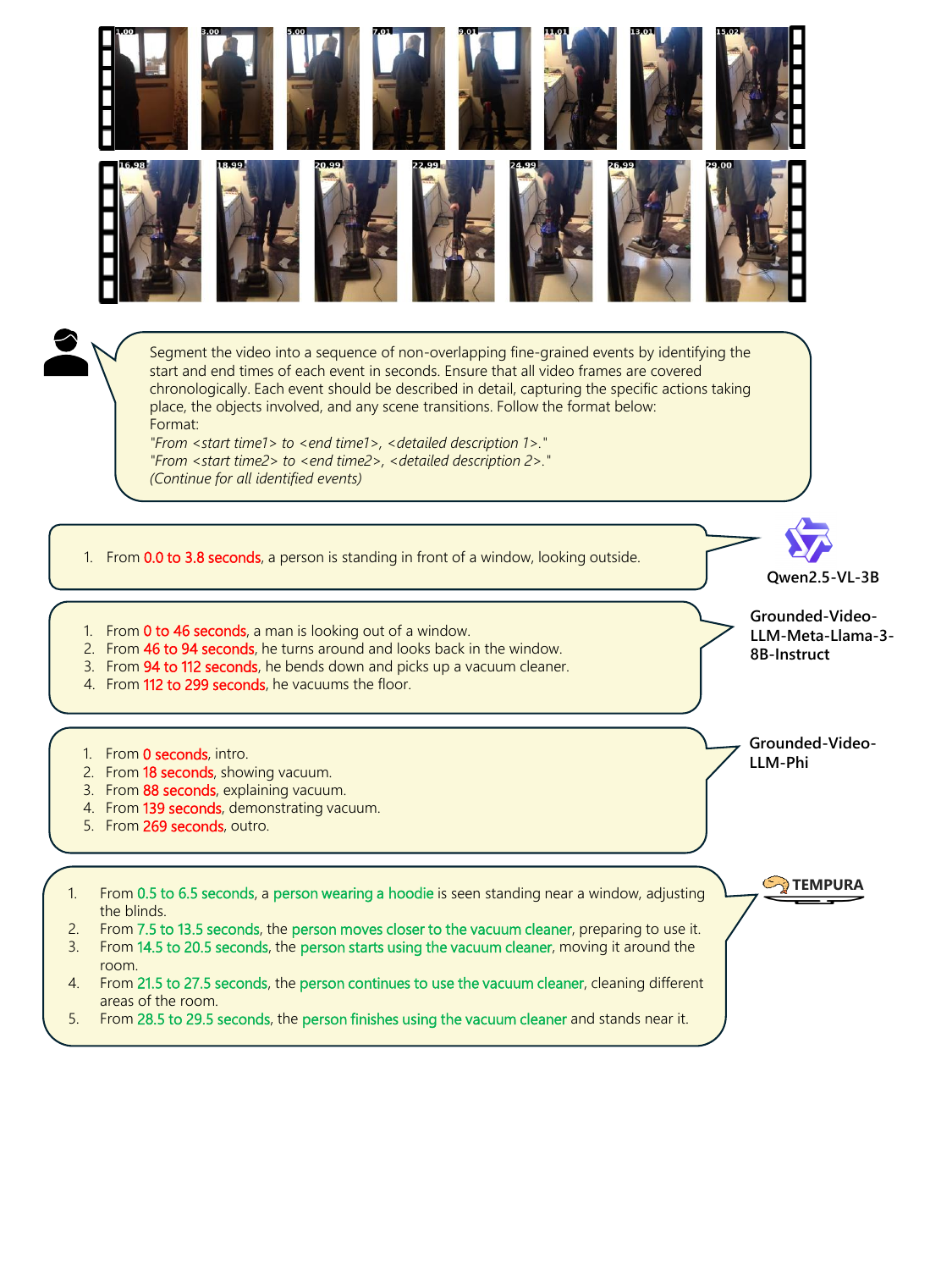}
    \caption{Comparing TEMPURA (our model) and other baseline model (Qwen2.5-VL-3B~\citep{qwenvl25}, Grounded-Video-LLM-Meta-Llama-3-8B-Instruct~\citep{wang2024groundedvideollmsharpeningfinegrainedtemporal}, and Grounded-Video-LLM-Phi~\citep{wang2024groundedvideollmsharpeningfinegrainedtemporal}) abilities to generate detailed descriptions on \textbf{fine-grained events on short videos}. Each model is prompted to segment the video into fine-grained events and describe the events in detail with correct start/end timestamps. Red text indicates incorrect responses with incorrect start/end timestamps or poor descriptions of the event segment.}
\label{fig:temp_rel4}
\end{figure*}

 Figure~\ref{fig:temp_rel4} compares the performance of Qwen2.5-VL-3B (our baseline), Grounded-Video-LLM-Meta-Llama-3-8B-Instruct~\citep{wang2024groundedvideollmsharpeningfinegrainedtemporal}, Grounded-Video-LLM-Phi~\citep{wang2024groundedvideollmsharpeningfinegrainedtemporal}, and TEMPURA (our proposed model) on fine-grained video segmentation.
 The red highlights indicate errors in timestamp predictions and the failure of other models to produce detailed event captions, even in shorter videos.
 In contrast, TEMPURA demonstrates better performance, as indicated by the green text, by producing more accurate timestamps, fine-grained events, and descriptive event captions.

\end{document}